\documentclass[fullpaper,cameraready]{nldl}

\renewcommand{\DOI}{10.7557/18.6807}

\usepackage[utf8]{inputenc}
\usepackage{url}
\usepackage{graphicx}
\usepackage{authblk}

\usepackage{algorithm}
\usepackage{algorithmic}
\usepackage{soulutf8}
\usepackage{svg}
\usepackage[compact]{titlesec}

\usepackage[square,numbers]{natbib}

\usepackage{hyperref}

\title{A Comparison Between Tsetlin Machines and Deep~Neural~Networks in the Context of Recommendation~Systems}
\author[1]{Karl Audun Kagnes Borgersen\thanks{Corresponding Author: karl.audun.borgersen@uia.no}}
\author[1]{Morten Goodwin}
\author[1]{Jivitesh Sharma}
\affil[1]{Centre for Artificial Intelligence Research, University of Agder}

\date{\vspace{-5ex}}

\begin{document}
\nldlmaketitle

\begin{abstract} 
Recommendation Systems (RSs) are ubiquitous in modern society and are one of the largest points of interaction between humans and AI. Modern RSs are often implemented using deep learning models, which are infamously difficult to interpret. This problem is particularly exasperated in the context of recommendation scenarios, as it erodes the user's trust in the RS. In contrast, the newly introduced Tsetlin Machines (TM) possess some valuable properties due to their inherent interpretability. TMs are still fairly young as a technology. As no RS has been developed for TMs before, it has become necessary to perform some preliminary research regarding the practicality of such a system. In this paper, we develop the first RS based on TMs to evaluate its practicality in this application domain. This paper compares the viability of TMs with other machine learning models prevalent in the field of RS. We train and investigate the performance of the TM compared with a vanilla feed-forward deep learning model. These comparisons are based on model performance, interpretability/explainability, and scalability. Further, we provide some benchmark performance comparisons to similar machine learning solutions relevant to RSs.
\end{abstract}


\section{Introduction}
Currently, most recommendation systems in production are either implemented with largely black-box models, such as deep neural networks (NN), proprietary systems, or through solutions utilizing overly simplistic collaborative filtering, which are incapable of processing arbitrary continuous and categorical features. Each of these causes difficulties when their developers wish to generate an explanation for a single prediction. Since black-box-like models are difficult to debug and modify, this often results in unfortunate complications, such as a music app presenting an inequitable presentation of artists based on nationality due to irrelevant common factors. Even though there exist tools to offset this problem, they are less effective than models that create these explanations innately \citep{interpretabilityexplainability}

Traditionally, explainability solutions for Machine Learning (ML) models determine their results based on modifying the model's input data, but such methods are severely hampered at larger scales. As the number of classes in a multi-class classification problem increases, so does the number of input values that need to be investigated. The reason is that these input values take the form of evaluating whether a change alters a single target output class. When the number of input features for a model decreases, so does the contribution of any single feature, making individual contributions more and more difficult to parse. Furthermore, these methods only elucidate correlations within the dataset. As laid out by \cite{interpretabilityexplainability}, these issues will persist while non-interpretable models remain in use. 

In this paper, we follow the definitions of \cite{interpretabilityexplainability} which separates between explainability and interpretability. Explainability, in this case, refers to modifications or tools applied to an ML model that is treated like a black box, i.e., we are unable to examine the model beyond its input and output. These model-agnostic methods are used to generate post-hoc explanations of ML models. Interpretability, in this case, is explanations generated as a by-product of the machine learning model. As outlined in \cite{interpretabilityexplainability}, explanations generated by explainable models can often be misleading and are therefore less desirable than explanations generated by fundamentally interpretable models. Explainable model tools detect patterns that are correlated with certain results, as opposed to interpretable models, which are capable of explaining the causation of certain results.

This paper presents a comparison between the TM and other widely-used explainability solutions. As such, we present experiments to compare the interpretability and explainability of the models, as well as to test and illustrate the viability of TM as an RS by comparing its performance to other ML models relevant within the field of RSs such as NNs,  Linear Regression (LR), Support Vector Machines (SVM), Decision Trees (DT), and Gradient Boosted Decision Trees (GBDT). These methods play an integral part in more advanced recommendation system solutions and will therefore give a reasonably good indication as to the viability of TMs for such roles. 
We evaluate the models according to (1) their performance, which means accuracy and mean precision at k, (2) interpretability and explainability, and (3) scalability. These evaluation metrics test the viability of the TM compared to more established models in terms of its strengths and weaknesses. This includes model performance to ensure competitive recommendation capabilities, interpretability, to highlight the differences in parsing recommendations between TM and other models. And scalability demonstrates whether the model can perform as well in more realistic conditions for an RS, meaning more data, more classes, and more stringent latency requirements.




\section{Related Work}
RSs are a useful tool for alleviating the problem of product variety information overload for a given user. \cite{rssurvey2022} is a recently published paper, provides an overview of the current state of the art. An inspiration for evaluating recommendation systems in this manner was a paper discussing the YouTube recommendation engine by \cite{youtuberec2016}. This paper discusses the use of a feed-forward architecture for both candidate generation and ranking.

A large amount of work has been dedicated to enhancing the explainability of machine learning models in general, and NNs in particular. In some cases, such as  Local Interpretable Model-Agnostic Explanations (LIME) by \cite{limepaper} or SHapley Additive exPlanations (SHAP) by \cite{shappaper}, this takes the form of model agnostic methods which investigate the impact of the model when certain input values are altered or omitted. This altered value is sampled from a distribution of training data. In the case of TransPer by \cite{karlsruheinterpretability} uses Layer-Wise Relevance Propagation to tailor interpretations to NNs.

There are existing interpretable machine learning solutions that are used in modern recommendation systems, such as gradient boosted trees \citep{gradientboostedtrees} and LR \citep{interpretlogisticregression}. LR tends to be lacking in performance, and gradient-boosted trees belong to a category of ML more performant in competitions rather than in research and industry. For instance, the winning solution in \cite{HnMKaggleDataset} heavily incorporates gradient tree boosting. \citep{competitionsnodl} discusses a few potential reasons why this might be the case.

Since the introduction of the TM in 2018 \cite{granmo2018tsetlin}, the TM has achieved competitive results in several fields. These include computer vision with the introduction of the convolutional TM \cite{convolutionaltm} outperforming 4-layer CNNs on the MNIST datasets, and natural language processing such as fake news detection by \cite{fakenewstmnlp} outperforming XLNet and BERT on the PolitiFact and GossipCop datasets.



\begin{figure*}[tbp]
    \centering
    \includegraphics[width=0.985\linewidth]{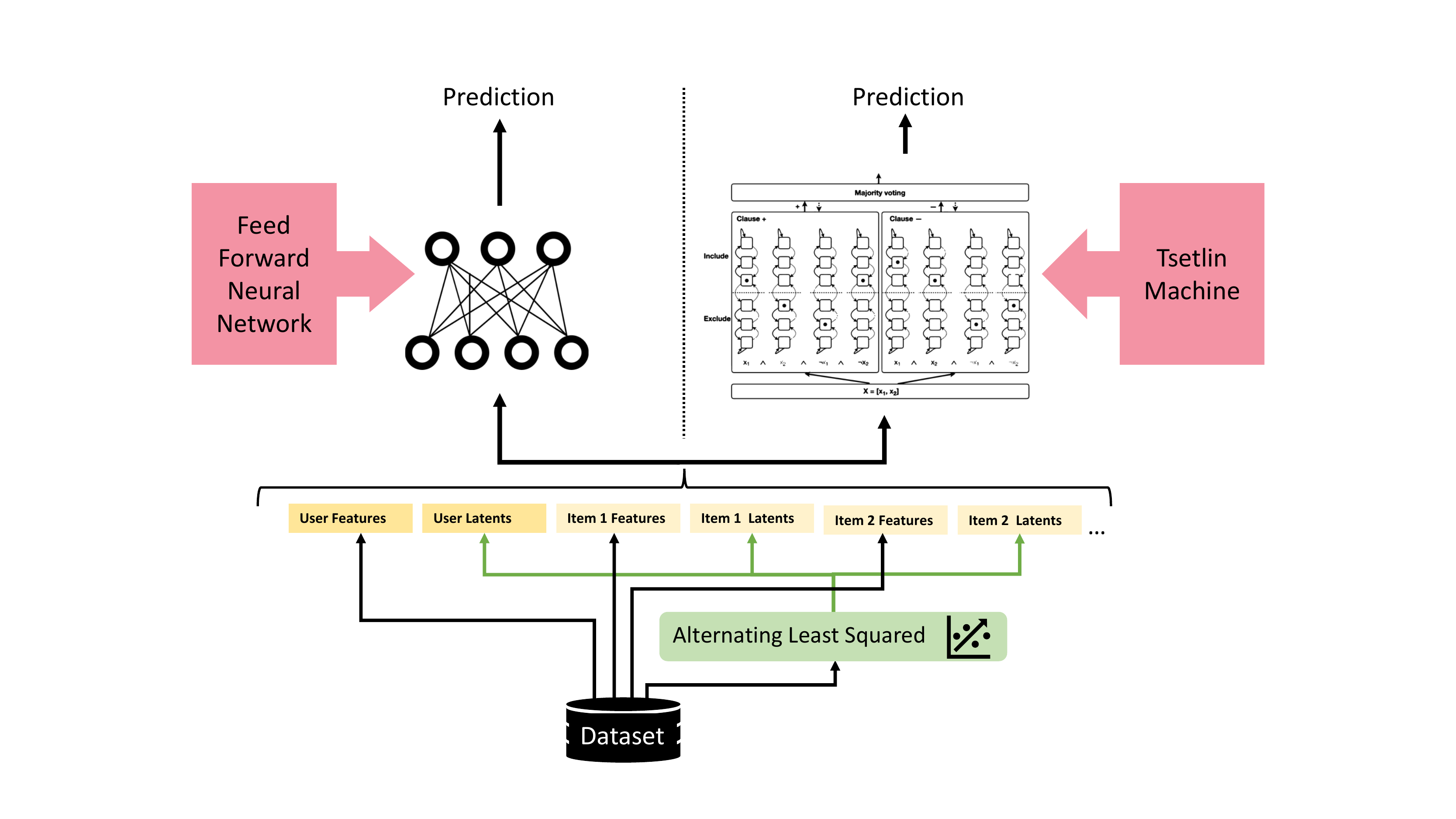}
    \caption{A Diagram of the project's architecture depicting the structuring of the dataset into input vectors and being fed through the NN and TM. The input has been structured into pair of attributes and ALS latents. With the first set of values denoting the customer and the final sequence denoting the attributes and latents of the customer's n most recent articles purchased.}
    \label{fig:modelarc}
    \vspace{-4mm}
\end{figure*}


\section{Method}
Figure \ref{fig:modelarc} shows a high-level overview of the TM and the NN\footnote{All code and experiments are openly available at \tiny{\url{https://github.com/cair/Tsetlin-Machine-Deep-Neural-Network-Recommendation-System-Comparison}}.}. We apply our models to the task of predicting customers' future purchases. The series of items purchased by the user is split based on whether the purchase was made during the last $30$ days, these subsets are used as training and testing data respectively. We make predictions by providing the models' input in the form of a set of customer and item features, by concatenating the relevant customer attributes with a series of item attributes. The items in this series are the attributes of the user's most recent $N$ purchases. If the purchase history does not extend to $N$ items, these are substituted with simple padding values. Further, to capture implicit relationships between customers and items, both are fed through Alternating Least Square matrix factorization as preprocessing. The resulting latents are used to  augment item and customer attributes.
Model performance is evaluated in an environment with $400$ classes. We choose $400$ classes as it represents a balance between an abnormally large count of classes and model performance with a limited amount of GPU memory available.

We compare the scalability of the different solutions by running these for $10$ epochs on differing-sized subsets of the dataset. Each subset contained the top $k$ most popular items from the dataset to ensure more viable runs in the lower item configurations. This weighting towards the use of popular items first meant that while the number of output items scaled consistently, the total portion of the dataset used did not. As there are far too many factors to balance for a fair comparison between the run times of the NN and the TM (e.g., model hyperparameters, system environment), the run times for each model are scaled relative to the run time of the lowest item configuration run. The width and height of the NN remained the same throughout all runs. Table \ref{tab:scalingdata} lists the relative sizes of the datasets for each run scale.

\begin{center}
\begin{table}[ht]
\begin{tabular}{||c | c c||} 
\hline
Num items & Item entries & Dataset percentage \\ [0.5ex] 
 \hline\hline
8 & 15208 & 1.616\% \\
\hline
64 & 75635 & 8.039\% \\
\hline
512 & 295944 & 31.456\% \\
\hline
2048 & 571843 & 60.781\% \\
\hline
all (28729) & 940816 & 100.000\% \\
\hline
\end{tabular}
\caption{Relative dataset scaling comparison}
\label{tab:scalingdata}
\end{table}
\vspace{-4mm}
\end{center}
%
%
\subsection{Dataset}
The dataset used for this project is constructed out of the dataset ``H\&M Personalized Fashion Recommendations" (H\&M dataset), which was used for a fashion-based recommendation system competition hosted on Kaggle by \cite{HnMKaggleDataset}. We chose this dataset because of its large size compared to other similar datasets and because it comes from a market leader. Thus, this grants us a larger set of diverse training data.
Further, the online leaderboard for the competition provides a good overview of how well a Mean Average Precision \@ 12 (MAP\@12) measures up to other existing solutions. Hence, we can easily compare our model with the published model results section of the H\&M Personalized Fashion Recommendations leaderboard. 
\subsection{Model setup}
Since we are more interested in discovering fewer sets of recurring patterns between user behavior and results, we reduce the number of TM clauses per class to far fewer than other state-of-the-art solutions. This was done primarily to ensure reasonable time and consumption per run but is unlikely to have impacted performance too badly. In contexts such as the MNIST dataset for convolutional TMs \citep{convolutionaltm}, we aim to discover a robust set of patterns that denote every single output class occurrence. In the MNIST example, $8000$ clauses are used per class, as compared to $200$ per class in this project. 

The NN solution consists of a series of five successive dense layers with sigmoid activations, culminating in a softmax activation. These layers have a gradually diminishing width of $[120, 110, 90, 80, 70]$.

Each of the other benchmark solutions is given the same input format as the NN.
\section{Results and Discussion}
\subsection{Model performance}
Table \ref{tab:modelperformance} compares the models' performance-based and Mean Precision at K (MAP@k). The results show a similar performance between the TM and NN. Both models fail to converge to a MAP@1 value, tending to oscillate by about 0.01\%. Even though all models perform significantly better than baseline calculated random guessing, i.e., simply selecting the most prevalent class from among the 8729 classes, they could yield even better results through hyperparameter tuning and scaling of the network. The results of the TM are likely hampered by the lack of integrated multi-label classification and significantly fewer clauses per class than most SoTA solutions. The NN model performs significantly worse than expected on larger map@k k values. This decreased performance is likely the result of a bug. Although not much time was spent experimenting with the GBDT solution, it would likely benefit from some hyperparameter optimization. Both the TM and NN significantly outperform the LR, SVM, and DT comparison methods and the GBDT solution to a lesser extent.

In terms of MAP@12, the TM solution outperforms the winning \cite{HnMKaggleDataset} solution of $0.03792$ by $0.007$. It should be noted that the TM has the significant advantage of heavily limiting possible output classes, from 8729 to 400. The reader should also note that the machine learning solutions presented operate with significant differences in both structures and hyperparameter setups, and hence, the comparability of performance could be improved. All in all, these model performance results should be considered promising. Though these results demonstrate the TM still slightly underperforms in terms of MAP@n. Despite the presence of highly continuous elements in the form of latent values, they still demonstrate the capability of detecting the highly sparse relationships inherent to RS problems at a comparable level to the NN.

\begin{center}
\begin{table}[!htb]
\begin{tabular}{||c | c c c||} 
\hline
Model & MAP@1 & MAP@12 & MAP@100 \\ [0.5ex] 
 \hline\hline
NN & \textbf{0.0219} & 0.0291 & 0.0409 \\
\hline
TM & 0.0215 & \textbf{0.0449} & \textbf{0.0554} \\
\hline
LR & 0.0138 & 0.0318 & 0.0410 \\
\hline
DT & 0.0152 & 0.0232 & 0.0232 \\
\hline
SVM & 0.0127 & 0.0303 & 0.0410 \\
\hline
GBDT & 0.0207 & 0.0368 & 0.0.0455 \\
\hline
\end{tabular}
\caption{Comparison of model}
\label{tab:modelperformance}
\vspace{-4mm}
\end{table}
\end{center}





\subsection{Interpretability and explainability}
This section discusses interpretability and explainability. We focus on the models' interpretability, i.e., parsing model predictions based solely on their internal workings, and the models explainability, i.e., parsing predictions through model agnostic methods.
\subsubsection{Interpreting the Tsetlin Machine}
\begin{figure}[h!tbp]
    \centering
    \includegraphics[width=\linewidth]{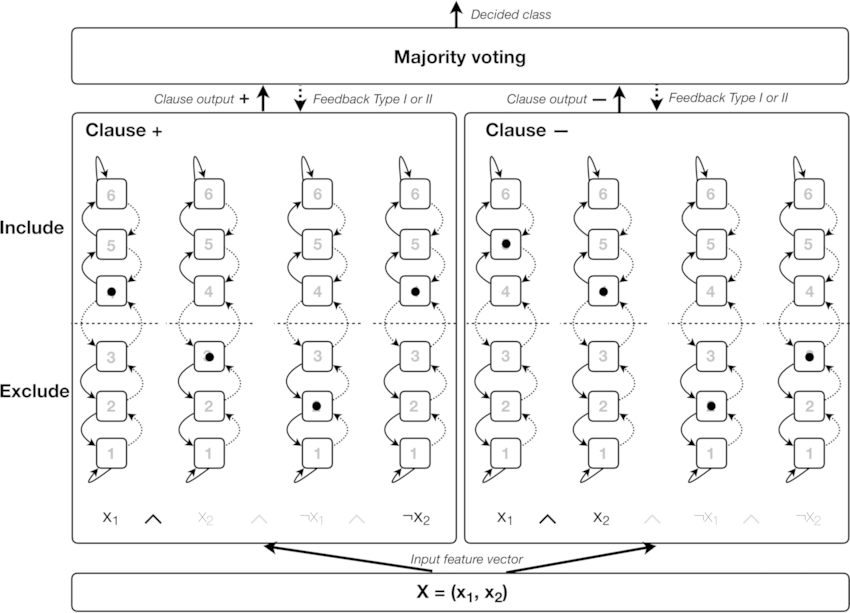}
    \caption{A general overview of the structure of TMs. Depicting a TM with two clauses and two features.\cite{granmo2018tsetlin}}
    \label{fig:tmdiagram}
\end{figure}

This paragraph briefly explains how the decision-making of a TM is structured. For further details on for instance, the feedback mechanism or how the TM maitains states, we refer the reader to \cite{tmbook}. An overview of a simplified TM is displayed in Figure \ref{fig:tmdiagram}. A TM consists of a series of Tsetlin Automata as the atomic component. Each of the classes in a TM form sets of Tsetlin Automata into clauses wherein each input feature is assigned two automata that determine whether a feature gets included or excluded. These clauses come in pairs of positive and negative clauses. All of these clauses vote independently to determine whether the input belongs to a class. A positive clause, e.g., $x1 \land \neg x2$, votes for this given class if feature $1$ equals $1$ and feature $2$ equals $0$. At the end of the voting round, the class with the highest total votes is predicted to be the class of the input features.

\begin{figure*}[h!tbp]
    \centering
    \includegraphics[width=\linewidth]{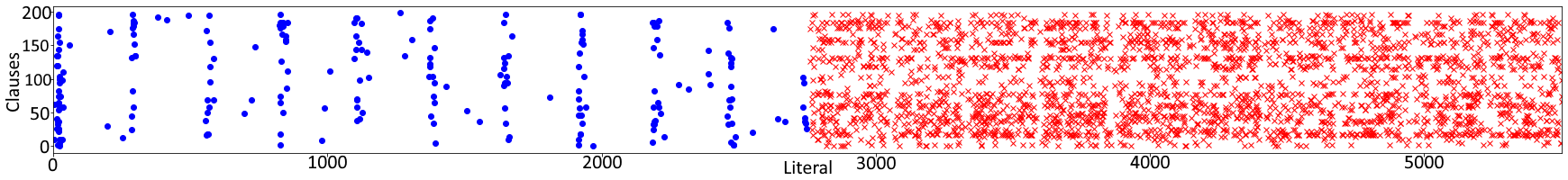}
    \caption{Visualisation of the TM clauses vs classes where the Y-axis denotes the clauses and X-axis denotes features for a single TM class. A blue circle is an included clause feature and a red cross is an excluded feature.}
    \label{fig:tmclause22}
\end{figure*}

Figure \ref{fig:tmclause22} presents each clause's set of included and excluded features. The graph is intended to give a rough overview of feature inclusion. For instance, we can observe set intervals of feature inclusion. These are correlated with purchased item attributes. Further, most clauses prioritize the initial features over later ones. These early features are related to user attributes.

The negative features are more difficult to parse due to their quantity. The prevalence of positive features compared to negative ones depends on prediction rates. Table \ref{tab:featurepresecense} shows the number of features present in clauses for a few example classes with differing prediction rates. In this case, count refers to how often the given class was predicted as the most likely class in the test dataset. The output classes with a larger presence in the dataset and within prediction show a correlation with more intricate positive patterns and less intricate negative clause patterns.
\begin{center}
\begin{table}[htb]
\begin{tabular}{||c | c c c c c||} 
\hline
Rarity & count & $pos$ & $\neg pos$ & $neg$ & $\neg neg$ \\ [0.5ex] 
 \hline\hline
prevalent & 5067 & 234 & 2917 & 105 & 92 \\ 
\hline
medium & 182 & 193 & 1243 & 119 & 79 \\ 
\hline
rare & 16 & 160 & 651 & 156 & 1093 \\ 
\hline
absent & 0 & 125 & 342 & 164 & 1374 \\
\hline
\end{tabular}
\caption{TM scaling clause complexity relative to prediction counts}
\label{tab:featurepresecense}
\vspace{-4mm}
\end{table}
\end{center}
By counting the occurrences of features in clauses, we can readily note outliers in terms of clause importance. $a0\_section\_name\_Collaborations$ is, for instance, the most heavily weighted feature, being included in $331$ positive clauses across all $400$ classes. For reference, the mean clause inclusion rate is $26.2$. Categorical values are included at a rate of $26.0$ as opposed to latent values, which are included at a rate of $27.7$. 

This section briefly discusses some TM global trends, but interpretation can be far more granular. For example, Class nr. $17$ has resulted in clauses that target young mothers purchasing items for their children. Class nr. $110$ targets similar items but includes some more upper body clothes as well. For the $249th$ user in our test dataset, these two classes are tied for being predicted as the most likely class. Both received a score of $92$. Had the 7th most recent purchase made had the spots graphical appearance tag, item nr. $110$ would have been the predicted item.
\subsubsection{Interpreting the Neural Network}
When referring to the NN so far, we've used the term `black-box-like' model. While it is not practically possible to read a NN's behavior based only on its weights and biases, due to the size and complex highly non-linear dependencies in the NN. The fact that the majority of the data is categorical further hinders our ability to interpret these weights. However, we can still observe some rudimentary behavior from the weights of the input layer. Note that all of the input features to the NN have been normalized when interpreting the model, which was done to make the first layer of weights and biases easier to parse. 

By summing the weight values from the input layer to the first hidden layer, we get some insight into the model's emphasis on each value. Customer age, for instance, is the only customer attribute with a positive weight sum. Customer factors, the latent features of each customer, heavily outweigh the other customer attributes by around a factor of 10. These weights indicate a higher relative importance for customer age and a significantly higher importance for customer factors than the other included features, which is in line with what we intuitively expect the NN to emphasize. 
Table \ref{tab:meanweightsnn} shows selected mean first layer weights.
\begin{center}
\begin{table}[!htb]
\begin{tabular}{||c c c||} 
\hline
club\_member\_status & age & user\_factor\_0 \\ [0.5ex] 
 \hline\hline
-0.098743 & 0.153792 & 0.976568 \\
\hline
\end{tabular}
\caption{Input weights for three NN input values.}
\label{tab:meanweightsnn}
\vspace{-4mm}
\end{table}
\end{center}
%
\subsubsection{Model Explainability}
\begin{figure}[h!tbp]
    \centering
    \includegraphics[width=\linewidth]{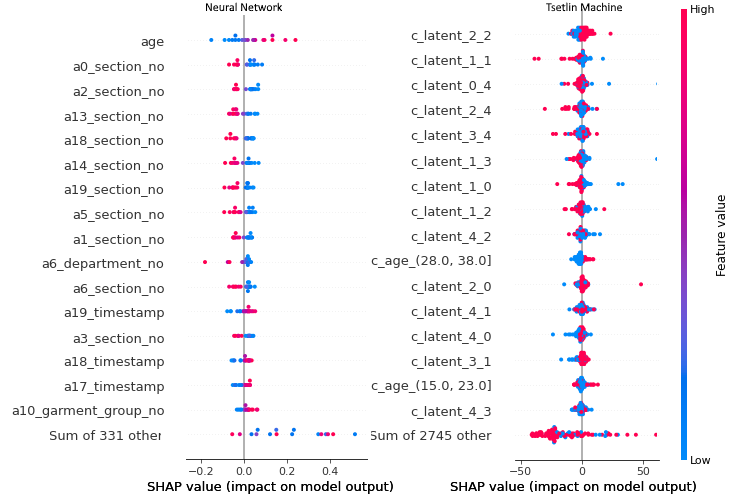}
    \caption{Beeswarm diagram of the top 10 most impactful features of a single class, using the NN model and the TM.}
    \label{fig:shapcombined}
\end{figure}

The previous section demonstrates the relative ease of interpreting a pure TM as compared to a pure NN. However, such comparisons do not tell the whole story. There exist a myriad of tools designed to bridge the gap between explainability and interpretability. For instance, the left of Figure \ref{fig:shapcombined} displays a SHAP beeswarm diagram of the $17$ most impactful features for a single class for the NN.


As SHAP is a model-agnostic tool, we can also use it to diagnose TM's predictions. The right side of Figure \ref{fig:shapcombined} displays the resulting SHAP values. Though these model impacts are a reflection of correlation on a subset of the data, as estimating SHAP values for every single test entry is too computationally and time intensive to be realistic, this approach is capable of pointing out which input elements caused a certain prediction from the TM. In this example the features with the decidedly most impact was the section numbers for the NN, as compared to the TM which had latent values as the most impactful features.

\subsubsection{Interpretability Performance}
The TM-based solution significantly outperforms NNs in interpretability. The entangling, non-linear dependencies of NNs render them essentially non-interpretable while the properties of the TM enable a degree of insight into RS behavior not present in the current SotA. As with most other machine learning models, the TM remains compatible with model-agnostic tools and should therefore be considered on par in terms of explainability.


\subsection{Scalability}
\begin{figure}[tbp]
    \centering
    \includegraphics[width=0.75\linewidth]{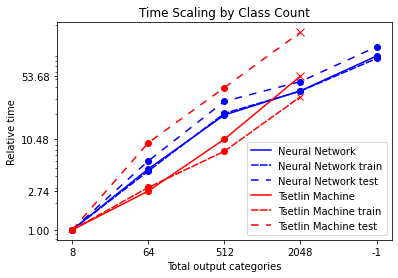}
    \caption{Time required to train the models for $10$ epochs, train the models for a single epoch, and test the models for a single epoch. The $X$ values denote the number of label classes present in the dataset, and the $Y$ values denote the time to train as compared to the smallest model configuration. e.g. the TM configuration with 64 items took $\sim1.98$ times longer to train than the smallest configuration of 8 items.}
    \label{fig:scalingbysize}
\end{figure}

All results for the TM described in this paper have been performed on the CUDA implementation of the TM. This implementation is remarkable for its almost constant time scaling in terms of the number of clauses \citep{desynctm}. However, the domain of RSs presents challenges in terms of the number of classes far beyond what the TM has been evaluated on before, and this section will therefore document the scaling potential of the TM as compared to the NN. Figure \ref{fig:scalingbysize} presents the relative scalability of the TM and NN. The NN training and run-time scaling performs worse with fewer classes but appears to handle extreme class quantities of $> 1000$ better than the TM. The TM performs consistently worse in terms of relative test time. Even in an environment with a generous portion of GPU memory available (up to 1500GB depending on server load), we could not evaluate the TM solution with all possible classes. These scalability issues will need to be resolved before a TM-based RS can be considered for more traditional RS problems.

\section{Future Work}
The structure of multi-class TMs causes a degree of modularity not present in NNs, ensuring that output classes can be inserted or removed without any modification to the rest of the model. Such a structure can be used to organize a far more dynamic training and evaluation process when compared to NNs. Essentially each class can be trained independently and with independent clauses. This opens the potential for a TM-based integrated candidate generation and ranking system in which the TM is tasked with performing multi-class classification on an entire inventory of items. We can calculate which clauses are described as the most prevalent when ranking each class/item. A potential approach to this is presented with the introduction of the integer weighted TM by \cite{intweightedtm}. We can then perform relatively cheap elimination of items that do not fulfill the highest weighted clauses for the given input. We then gradually apply more and more clauses for the remaining items/categories until a final selection of top items remains.\\
\section{Conclusion}
While the TM-based recommendation system solution presented in this paper is not viable for use in a production environment, we have presented a direct performance comparison to existing RS methods. The TM is capable of competing with the NN in terms of performance while providing a significant improvement in interpretability. However, the TM's scaling capabilities show some worrying trends in terms of both time and memory consumption. Finally, we have proposed a potential architecture for overcoming some of these issues and playing to the TM's strengths.


\bibliographystyle{abbrvnat}
\bibliography{references}

\end{document}